\documentclass[sigconf]{acmart}
\usepackage{mdwmath}
\usepackage{mdwtab}
\usepackage{booktabs}

\usepackage{amsfonts}
\usepackage{multirow}
\def\ie{\emph{i.e.}}
\def\eg{\emph{e.g.}}

\newcommand{\bm}[1]{\mbox{\boldmath{$#1$}}}

\newcommand{\equref}[1]{(\ref{#1})}

\newcommand{\myPara}[1]{\vspace{.05in}\noindent\textbf{#1}}

\newcommand{\bl}[1]{\textbf{#1}}
\newcommand{\ul}[1]{\underline{#1}}
\newcommand{\mc}[1]{\mathcal{#1}}
\newcommand{\mb}[1]{\mathbb{#1}}

\begin{document}
\title{Collaborative Annotation of Semantic Objects in Images with Multi-granularity Supervisions}

\author{Lishi Zhang, Chenghan Fu, Jia Li$^*$}
\affiliation{\institution{State Key Laboratory of Virtual Reality Technology and Systems, SCSE, Beihang University}}
\email{jiali@buaa.edu.cn}

\begin{abstract}
Per-pixel masks of semantic objects are very useful in many applications, which, however, are tedious to be annotated. In this paper, we propose a human-agent collaborative annotation approach that can efficiently generate per-pixel masks of semantic objects in tagged images with multi-granularity supervisions. Given a set of tagged image, a computer agent is first dynamically generated to roughly localize the semantic objects described by the tag. The agent first extracts massive object proposals from an image and then infer the tag-related ones under the weak and strong supervisions from linguistically and visually similar images and previously annotated object masks. By representing such supervisions by over-complete dictionaries, the tag-related object proposals can pop-out according to their sparse coding length, which are then converted to superpixels with binary labels. After that, human annotators participate in the annotation process by flipping labels and dividing superpixels with mouse clicks, which are used as click supervisions that teach the agent to recover false positives/negatives in processing images with the same tags. Experimental results show that our approach can facilitate the annotation process and generate object masks that are highly consistent with those generated by the LabelMe toolbox.
\end{abstract}


%  \caption{This is a teaser}

\maketitle

\let\thefootnote\relax\footnotetext{*Corresponding author: Jia Li}

\section{Introduction}

In the past decade, large-scale datasets have greatly boosted the development of computer vision techniques. Typically, images in such datasets are annotated with one or several tags to depict the semantic categories of primary objects. In recent applications such as autonomous driving~\cite{chen2014beat,mousavian20173d,geiger2012we,chen2015deepdriving}, robot navigation~\cite{chen2017only,kim2017direct,mao2017shape,fankhauser2015kinect,zhu2017target} and visual question answering~\cite{goyal2017making,xu2017video,zhao2017video,zhou2017more}, however, knowing only the \textit{what} attributes represented by such tags becomes insufficient. To interact with the real-world scenarios, these applications often have to train and test models based on per-pixel masks of semantic objects, within which both the \textit{what} and \textit{where} attributes are simultaneously annotated.

\begin{figure}[t]
\centering
\includegraphics[width=1.00\columnwidth]{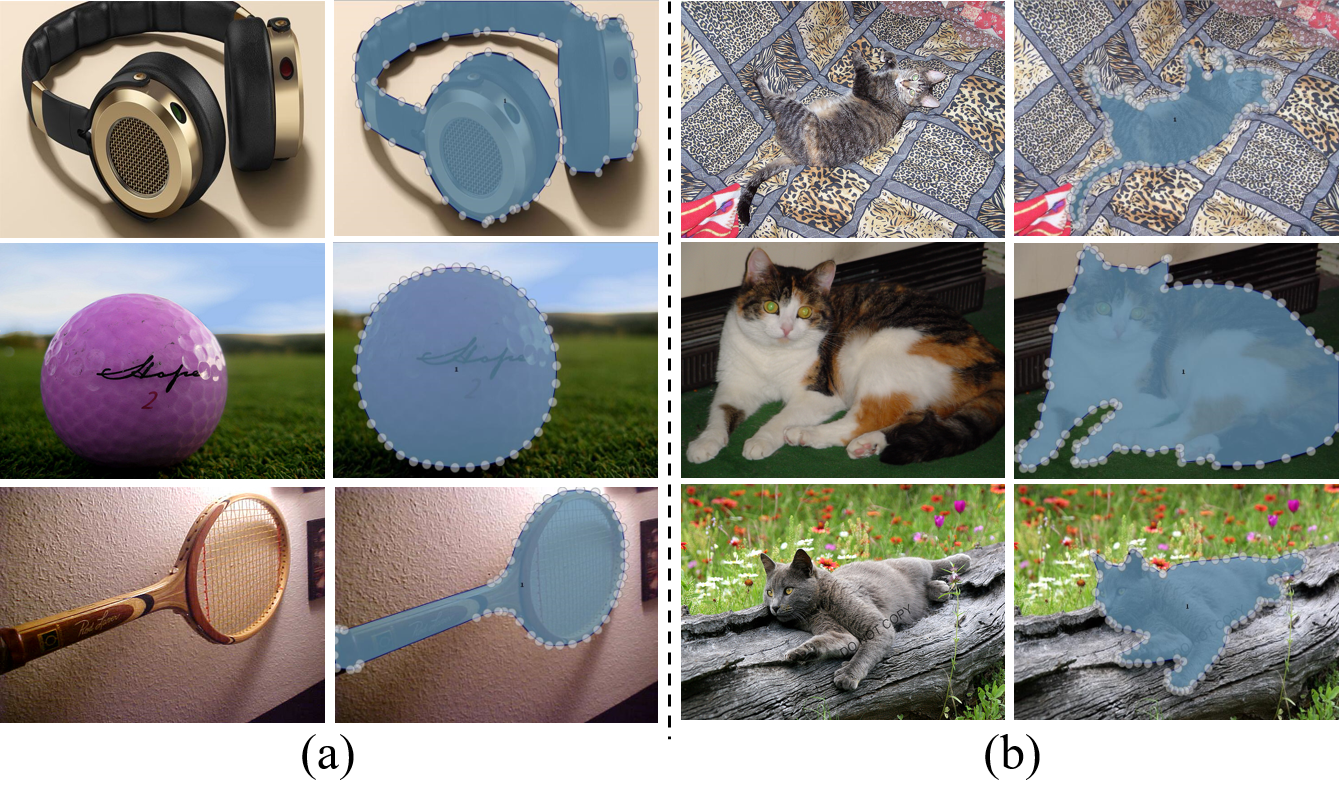}
\caption{Two time-consuming tasks in annotating semantic objects. a)~Approximating obvious non-polygonal boundaries with polygons; b)~Separating ambiguous boundaries where targets and distractors have similar visual appearance. In most cases, a computer agent performs good at the former task, and the human annotator is skilled in the latter task. This motivates the proposal of our human-agent collaborative annotation approach.}
\label{fig:twotask}
\end{figure}

Regardless of the overwhelming requirement of per-pixel masks of semantic objects, unfortunately, the pixel-wise annotation is often inefficient and tedious. It is difficult to construct datasets with per-pixel masks ~\cite{cordts2016cityscapes,gould2012multiclass} of semantic objects that are as large as existing datasets of tagged images. For example, ImageNet~\cite{deng2009imagenet,russakovsky2015imagenet} contains 14.2 million tagged images from 21,841 object categories, while MS COCO~\cite{lin2014microsoft}, a large dataset with per-pixel masks of 80 object categories, contains only 328k images. As shown in Fig.~\ref{fig:twotask}, two of the most time-consuming tasks in annotating semantic objects include 1)~approximating obvious non-polygonal boundaries by drawing polygons and 2) separating ambiguous boundaries where targets and distractors have similar visual appearance. In most cases, a computer agent performs good at the former task, and the human annotator is skilled in the latter one. Therefore, it is necessary to develop a human-agent collaborative annotation approach to speed up the annotation process so that large-scale datasets with per-pixel masks of semantic objects can be efficiently generated.

\begin{figure*}[t]
\centering
\includegraphics[width=1.00\textwidth]{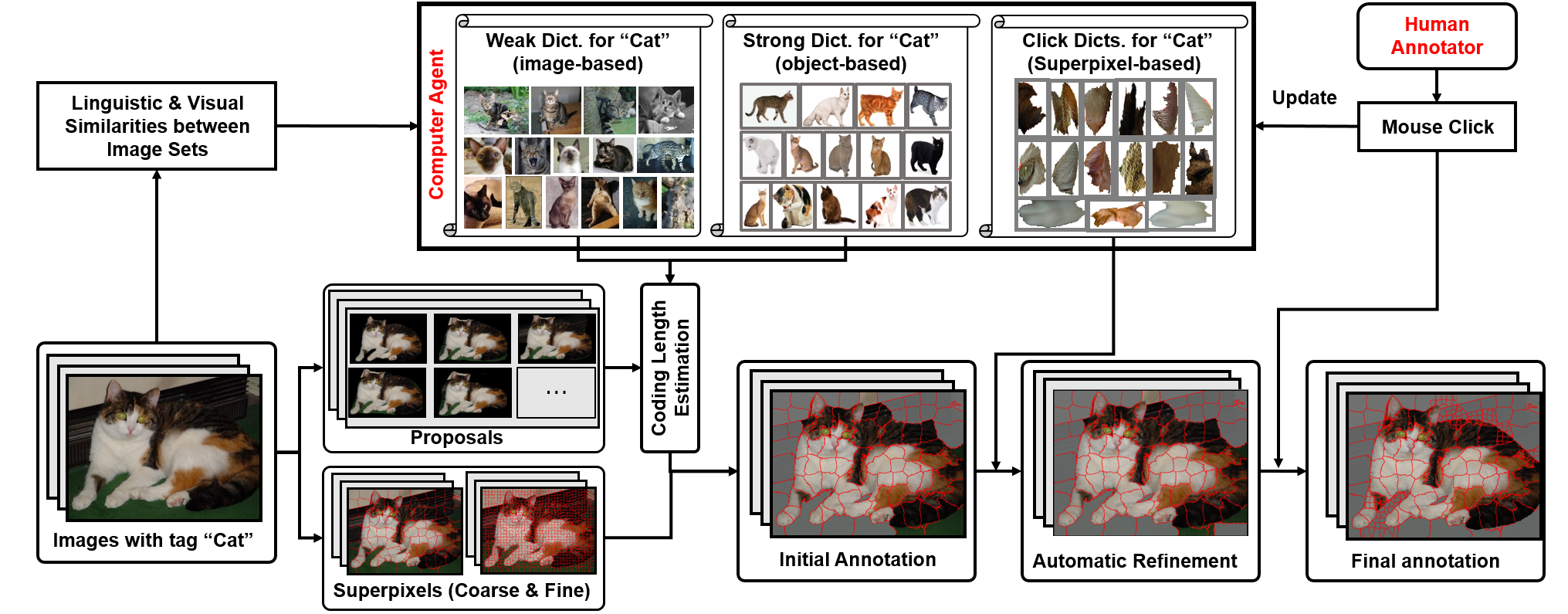}
\caption{Framework of the proposed approach. Given a set of images tagged with ``Cat,'' a computer agent is dynamically generated with weak, strong and click dictionaries. It first extracts object proposals and superpixels, and the tag-related object proposals are then inferred by measuring the sparse coding length of weak and strong dictionaries. By converting the tag-related objects into the binary labels of superpixels, the human annotator can participate to flip the superpixel label or divide coarse superpixel into finer ones via mouse clicks. Such clicks are then used to form flip dictionaries which can be used to supervise the automatic refinement of subsequent images.}
\label{fig:framework}
\end{figure*}

The problem of human-agent collaborative annotation~\cite{su2010web} has been studied for many years. Based on the collaboration strategy, existing approaches can be divided into two major categories, which can be denoted as \textit{agent-decision} and \textit{human-decision}, respectively. Approaches in the agent-decision group require human interactions like scribbles~\cite{russell2008labelme} and bounding boxes~\cite{dai2015boxsup} at the beginning of the annotation process, followed by the automatic boundary refinement algorithms such as GrabCut~\cite{rother2004grabcut,chen2014beat} or Convolutional Neural Network (CNN)~\cite{lin2016scribblesup,uhrig2016pixel,castrejon2017annotating} that can be viewed as an agent. Such human-agent interactions can be repeated several times until satisfactory results are generated. A major drawback of the agent-decision approaches is that the final decision is made by the agent. Since the correlation between the input human interactions and the output annotation results is not very intuitive, these annotation results are rarely used as the official ground-truth.

For the human-decision group, the annotation process starts with an agent that automatically generates coarse annotation results in terms of superpixels~\cite{li2016iterative,shi2017weakly}, bounding boxes~\cite{kuettel2012segmentation,jain2016active,su2012crowdsourcing} or polygons~\cite{castrejon2017annotating}. These results are then manually refined by the human annotators through dragging and dropping anchor points~\cite{castrejon2017annotating} or clicking superpixels~\cite{kim2011markup}. Since the human annotator in these approaches makes the final decision, the generated masks can be used as the official ground-truth. Typically, the more powerful an agent is, the more time cost can be saved. However, one drawback of existing human-decision approaches is that the agents are often static with pre-trained (or pre-defined) parameters, while in annotating certain unusual categories of objects there may lack sufficient data for training such an agent. Moreover, the agents may repeatedly make the same mistakes even in segmenting the same categories of semantic objects. Therefore, it is necessary to develop a human-agent collaborative annotation approach, in which the agent is dynamically formed and gradually evolves to make less mistakes, and the human makes the final decision.

Toward this end, this paper proposes a human-agent collaborative approach that can efficiently generate per-pixel masks for tagged images. The main objective is to rapidly convert tagged images in previous datasets (\eg, ImageNet) into per-pixel masks of semantic objects. In this manner, a large-scale image dataset with per-pixel masks of semantic objects can be efficiently created. The system framework of the proposed approach is shown in Fig.~\ref{fig:framework}, which takes a set of images with the same tags as the input. In this framework, a dynamic agent is first created from tagged images and previously annotated semantic objects that have high linguistic and visual similarities with the input images. After that, a weak dictionary is constructed from images with similar tags, and a strong dictionary is constructed from previously annotated semantic objects. Based on the weak and strong supervisions from the two dictionaries, the agent first extracts a set of object proposals from each image and then measures their relationship with image tags according to the sparse coding length ~\cite{lee2007efficient}(\ie, the difficulty of encoding the features of object proposals from the two dictionaries). Such object-level relationships are then used to initialize the binary labels of coarse superpixels, based on which a human annotator interacts with the superpixels with mouse clicks. In the human annotation stage, simple mouse clicks are used to invert the label of a superpixel and divide a large superpixel into much smaller ones that are better aligned with ambiguous object boundaries. Meanwhile, superpixels whose labels are corrected by the annotators are recorded to update a flip dictionary which records false positives/negatives. In annotating subsequent images with the same tags, the initialization results can be automatically refined by the agent that gradually learns how to recover false positives/negatives from the flip dictionary. Experimental results show that our approach can facilitate the annotation process of semantic objects, and the collaborative masks have 91.21\% agreement with the masks generated by the LabelMe toolbox~\cite{russell2008labelme} in terms of F-Measure.

The main contributions of this paper include: 1)~We propose a human-agent collaborative approach to convert tagged images in previous datasets into per-pixel masks of semantic objects; 2)~We create a dynamic agent that can gradually evolve in facilitating the annotation process; and 3)~We conduct extensive experiments which not only validates the effectiveness of the proposed approach in speed up the annotation but also show that the collaborative annotation results have high agreement with the masks generated by the LabelMe toolbox.

\section{Automatic Annotation Initialization with Weak and Strong Supervisions}
The main objective of this work is to develop a collaborative tool that can rapidly convert the semantic tags of images in previous large-scale datasets into per-pixel masks of semantic objects. Without loss of generality, we assume there exists an image dataset in which only the primary object(s) of each image is annotated by several nouns (like ImageNet). Given a set of images annotated by the same set of tags, we first retrieve images and previously annotated object masks with relevant tags to construct two dictionaries, which are then used as the weak and strong supervision to automatically initialize the masks of semantic objects.

\subsection{Weak and Strong Dictionaries}
Let $\{\mb{I}_i,\forall i\}$ be all image sets and $\mb{W}_{i}$ be the set of tags of $\mb{I}_i$. The similarity between $\mb{I}_{i}$ and $\mb{I}_{j}$ is defined as
\begin{equation}\label{eq:sim}
S(\mb{I}_{i},\mb{I}_{j})=\frac{2\cdot{}S_l(\mb{W}_{i},\mb{W}_{j})\cdot{}S_v(\mb{I}_{i},\mb{I}_{j})}{S_l(\mb{W}_{i},\mb{W}_{j})+S_v(\mb{I}_{i},\mb{I}_{j})},
\end{equation}
where $S_l(\mb{W}_{i},\mb{W}_{j})$ and $S_v(\mb{I}_{i},\mb{I}_{j})$ denote the linguistic and visual similarities, respectively. The linguistic similarity is computed as
\begin{equation}\label{eq:lingSim}
S_l(\mb{W}_{i},\mb{W}_{j})=\frac{\sum_{\mc{W}_1\in\mb{W}_i}\sum_{\mc{W}_2\in\mb{W}_j}\bl{u}_1^{\text{T}}\bl{u}_2}{|\mb{W}_i|\cdot|\mb{W}_j|},
\end{equation}
where $\mc{W}_1$ and $\mc{W}_2$ are two tags represented by two normalized feature vectors $\bl{u}_1$ and $\bl{u}_2$, respectively. Here $\bl{u}_1$ and $\bl{u}_2$ are formed by using the pre-trained word2vec model~\cite{mikolov2013efficient,kim2014convolutional}, and their similarity is computed as the Cosine distance. $|\mb{W}_i|$ and $|\mb{W}_j|$ denote the numbers of tags in $\mb{W}_i$ and $\mb{W}_j$, respectively.

To compute the visual similarity, we adopt the DPN model \cite{chen2017dual}, a pre-trained Convolutional Neural Network, to convert each image in $\mb{I}_{i}$ and $\mb{I}_{j}$ into a 2688D feature vector. These vectors are also normalized to unit vectors, and the visual similarity is computed as
\begin{equation}\label{eq:visSim}
S_v(\mb{I}_{i},\mb{I}_{j})=\frac{\sum_{\mc{I}_1\in\mb{I}_i}\sum_{\mc{I}_2\in\mb{I}_j}\bl{v}_1^{\text{T}}\bl{v}_2}{|\mb{I}_i|\cdot|\mb{I}_j|},
\end{equation}
where $\mc{I}_1$ and $\mc{I}_2$ are two images represented by feature vectors $\bl{v}_1$ and $\bl{v}_2$, respectively. $|\mb{I}_i|$ and $|\mb{I}_j|$ denote the numbers of images in $\mb{I}_i$ and $\mb{I}_j$, respectively.

By incorporating the linguistic similarity \equref{eq:lingSim} and visual similarity \equref{eq:visSim} into \equref{eq:sim}, we can create a ranked list of image sets in decreasing order of the overall similarity. From this list, we select the top-ranked image sets with the similarity value above $0.95$. We assume that the per-pixel masks of some selected image sets are already annotated, while the others are not. As a result, we can construct a weak dictionary $\bl{D}_i^{w}$ and a strong dictionary $\bl{D}_i^{s}$ for these two groups of image sets. The weak dictionary is represented by a matrix whose columns are the feature vectors of the selected images, and the per-pixel masks are not annotated yet. The strong dictionary is constructed similarly on previously annotated objects by setting the image pixels outside the per-pixel masks to zero in extracting the object-based feature descriptors. The difference between the two dictionaries is that, features in the strong dictionary acts more precise in representing the attributes of semantic objects. To speed up the computation, we compress the 2688D feature vectors into 100D via Principal Component Analysis in constructing the two dictionaries.
\begin{figure*}[t]
\centering
\includegraphics[width=1.00\textwidth]{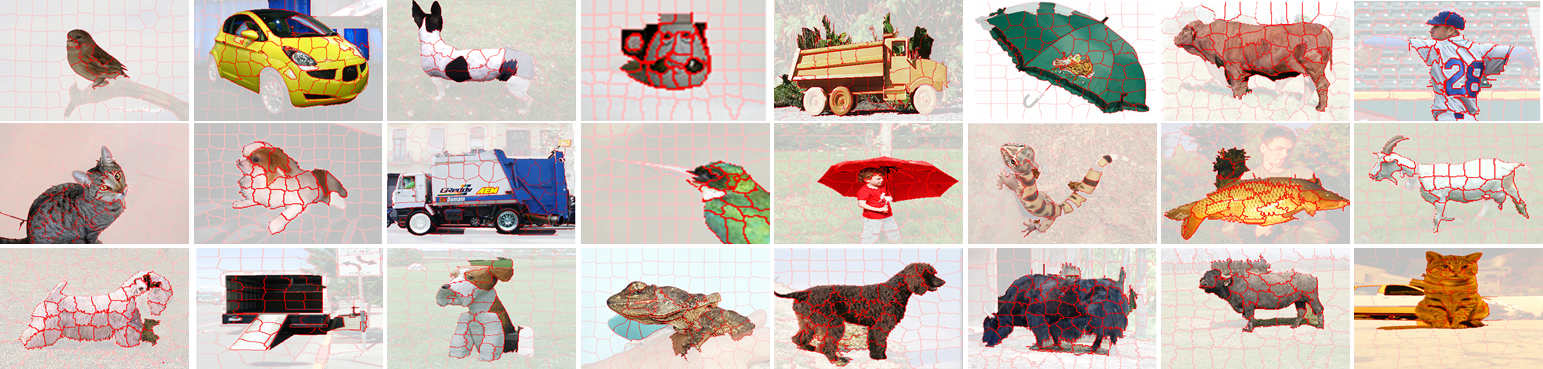}
\caption{Representative examples of initialization with weak and strong supervisions. We can see that the majority of semantic objects can be successfully detected, while some superpixels near ambiguous boundaries and in object-like regions may be assigned with wrong labels or have inappropriate scales. }
\label{fig:initial}
\end{figure*}
\subsection{Object Mask Initialization with Weak and Strong Supervisions}
Given the two dictionaries $\bl{D}_i^{w}$ and $\bl{D}_i^{s}$ that are related to the image set $\mb{I}_i$, we can automatically initialize object masks for all images in $\mb{I}_i$. For an image $\mc{I}\in\mb{I}_i$, we first use the MCG model ~\cite{arbelaez2014multiscale} to extract a number of object proposals and select the proposals with the highest objectness for subsequent analysis. Let $\mb{C}=\{\mc{C}_t\}_{t=1}^{|\mb{C}|}$ be the selected object proposals (we empirically set $|\mb{C}|$=200), we extract their visual features by using the pre-trained DPN model ~\cite{chen2017dual} and compress them into 100D descriptors by using Principal Component Analysis. Such compressed descriptors are denoted as $\{\bl{x}_t\}_{t=1}^{|\mb{C}|}$. For each proposal $\mc{C}_t\in\mb{C}$, its probability of being related to image tags can be computed according to the difficulties of encoding the proposal's features with the weak and strong dictionaries. That is, the easier a proposal can be encoded from the images and objects of the same category, the more likely it belongs to the desired sematic object. The sparse representation of the feature vector $\bl{x}_t$ can be estimated by solving
\begin{equation}\label{eq:sparseCoding1}
\begin{split}
{\bm \alpha}_{t}^{w}=\arg\min_{\bm \alpha}|{\bm \alpha}|_0,~ ~s.t.~\|\bl{x}_{t}-\bl{D}_i^{w}{\bm \alpha}\|_2^2\leq{}\epsilon,\\
{\bm \alpha}_{t}^{s}=\arg\min_{\bm \alpha}|{\bm \alpha}|_0,~ ~s.t.~\|\bl{x}_{t}-\bl{D}_i^{s}{\bm \alpha}\|_2^2\leq{}\epsilon,
\end{split}
\end{equation}
where ${\bm \alpha}_{t}^{w}$ and ${\bm \alpha}_{t}^{s}$ are the sparse representations of the input feature vector $\bl{x}_t$. The optimization objective of \equref{eq:sparseCoding1} is to minimize the two coding lengths of a feature vector based on the two dictionaries, which can be solved with the Orthogonal Matching Pursuit algorithm~\cite{donoho2012sparse}. The constant $\epsilon$ is a predefined threshold and we empirically set $\epsilon=0.1$ in all experiments.

Given the two sparse representations of an object proposal, its coding length can be estimated as
\begin{equation}
L(\mc{C}_t)=\min(|{\bm \alpha}_t^w|_0,|{\bm \alpha}_t^s|_0),
\end{equation}
Note that we only use the minimum coding length to measure difficulty of reconstructing the object proposals from tagged images and clean objects. With the coding length, the probability that $\mc{C}_t$ is related to the semantic object $\mc{O}$ described by image tags, denoted as $P(\mc{C}_t\in\mc{O})$, can be heuristically estimated as
\begin{equation}
\begin{split}
P(\mc{C}_t\in\mc{O})=\left(\frac{\max\{L(\mc{C}_t),\forall t\}-L(\mc{C}_t)}{\max\{L(\mc{C}_t),\forall t\}-\min\{L(\mc{C}_t),\forall t\}}\right)^Q
\end{split}
\end{equation}
%,\forall \mc{C}\in\mb{C}
where $Q$ is a constant that is used to suppress superpixels with large coding lengths (we empirically set $Q$=2).

After estimating such probabilities of all object proposals, we convert them into binary labels of superpixels for subsequent human interactions, and the annotation process can be formulated as finding a subset of superpixels that 1)~their union perfectly covers the semantic objects, and 2)~their number is small to reduce the time cost in human decision. Toward this end, we first extract a set of non-overlapping superpixels ~\cite{achanta2017superpixels}at a coarse scale, denoted as $\mb{S}$. The probability that a coarse superpixel $\mc{S}\in\mb{S}$ is related to image tags can be computed as
\begin{equation}
\mc{P}(\mc{S}\in\mc{O})=\frac{\sum_{p\in{}\mc{S}}\sum_{t=1}^{|\mb{C}|}\delta(p\in{}\mc{C}_t)\cdot{}P(\mc{C}_t\in\mc{O})}{\sum_{p\in{}\mc{S}}\sum_{t=1}^{|\mb{C}|}\delta(p\in{}\mc{C}_t)}
\end{equation}
where $p$ is a pixel and $\delta(p\in{}\mc{C}_t)$ is an indicator function that equals 1 if $p\in{}\mc{C}_t$ holds and 0 otherwise. As a result, the binary label of a superpixel $\mc{S}\in\mb{S}$, denoted as $G(\mc{S})$, can be initialized with a predefined threshold
\begin{equation}
G(\mc{S})=\left\{
\begin{array}{rcl}
1     &      & \mc{P}(\mc{S}\in\mc{O}) \geq{} \beta_0\\
0     &      & otherwise
\end{array} \right.
\end{equation}
where $\beta_0$ is a predefined threshold which is empirically set to 0.4. As shown in Fig.~\ref{fig:initial}, the initialized annotations can roughly cover the majority of primary objects, but may fail at ambiguous boundaries and certain object-like regions. Therefore, such superpixels should be refined by removing false positives, recovering false negatives and splitting into smaller superpixels.

\section{Human-Agent Collaborative Refinement with Click Supervision}
Based on the object masks initialized by the agent, a human annotator is involved for the collaborative annotation refinement. Such refinement can take two forms in processing the image set $\mb{I}_i$. For the very first images, only the manual refinement is used, and the revisions made by the annotator are recorded to create and update one additional flip dictionary. After that, the flip dictionary is used to provide automatic refinement before the manual stage to prevent the agent from making the same mistakes when processing the subsequent images.

\subsection{The Flip Dictionary From Clicks}
To speed up the manual refinement, we divide an image into massive superpixels at a much finer scale and further divide such superpixels by the boundaries of coarse superpixels. As a result, we can safely assume that superpixels at this scale align well with object boundaries. Based on the coarse and fine superpixels, only two click actions are require to refine object masks: 1)~a left click flips the superpixel label between 0 and 1; and 2)~a right click converts a coarse superpixel into much finer ones with the same labels. With these clicks, we can efficiently generate a per-pixel mask for an image. Compared with dragging anchor points to form polygons, the superpixel-based interactions are often easier to be conducted and can make use of the impressive capability of agents in segmenting non-polygonal boundaries.

\begin{figure}[t]
\centering
\includegraphics[width=1.00\columnwidth]{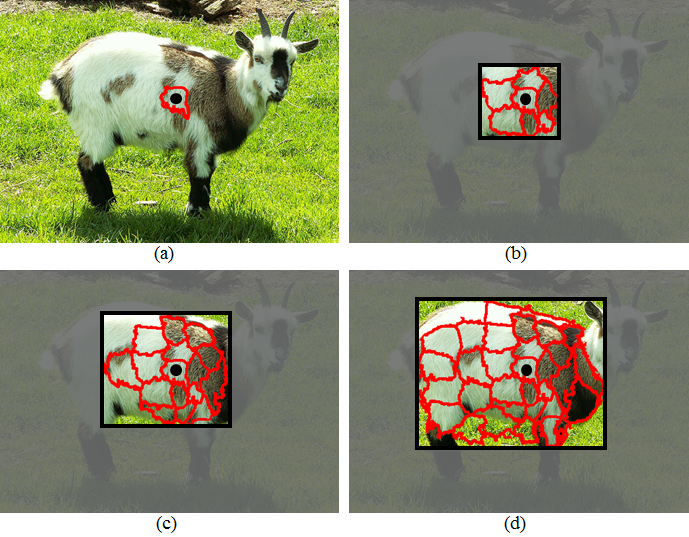}
\caption{A superpixel is embedded into multiple local contexts for multi-scale visual feature extraction. (a) A coarse superpixel from an input image (marked with the red contour and a black dot); (b)-(d) the local contexts (bounding boxes) with order-$1$, order-$2$ and order-$3$ neighbors. We can see that by embedding the superpixel into multiple contexts, its visual attributes can be described from both the local, regional and global perspectives.}
\label{fig:context}
\end{figure}

Meanwhile, for a coarse superpixel $\mc{S}\in\mb{S}$, we embedded source image into multiple local contexts at $N$ scales. Here the local context at the scale $n$ is represented by a bounding box covering the superpixel and all its order-$n$ neighbors. As shown in Fig.~\ref{fig:context}, the order-$n$ neighbors are adjacent to the order $n-1$ neighbors, and the order-$1$ neighbors are adjacent to $\mc{S}$. By embedding a superpixel into multiple local contexts, we can extract its visual features at $N$ scales by DPN model \cite{chen2017dual}, denoted as $\{\bl{y}_n,n=1,\ldots,N\}$. Then, for the coarse superpixels that are initialized with labels of 1 and receive left clicks, we use their features extracted at $N$ scales to form a flip dictionary $\bl{D}_i^{f+}$ that are used to record false positives. Similarly, a flip dictionary $\bl{D}_i^{f-}$ is also formed to record false negatives.

%
%\begin{figure}[t]
%\centering
%\includegraphics[width=1.00\columnwidth,height=5.0cm]{empty.png}
%\caption{Representative examples of annotation results at different stages.}
%\label{fig:repResult}
%\end{figure}

\subsection{Automatic Annotation Refinement Supervised by The Flip Dictionary}
The flip dictionary record the failures made by the agent in annotating the first several images from an input image set. In annotating subsequent images, we expect that the agent can gradually evolve by learning from such failures on how to recover false positives/negatives. In practice, we focus on the coarse superpixel near the classification boundaries (\ie, $\beta_0-\Delta\beta\leq{}\mc{P}(\mc{S}\in\mc{O})\leq{}\beta_0+\Delta\beta$, where $\Delta\beta$ is empirically set to 0.15). For such uncertain coarse superpixels with initial labels of 1, we first adopt the same sparse coding scheme to recognize probable false positives by solving
\begin{equation}
{\bm \alpha}_{n}^{f+}=\arg\min_{\bm \alpha}|{\bm \alpha}|_0,~ ~s.t.~\|\bl{y}_{n}-\bl{D}_i^{f+}{\bm \alpha}\|_2^2\leq{}\epsilon,
\end{equation}
After that, we refer to the coding length of ${\bm \alpha}_{n}^{f+}$ to selectively invert the labels of probable false positives:
\begin{equation}
G^*(\mc{C})=\left\{
\begin{array}{rcl}
1-G(\mc{C})       &      & \min\{|{\bm \alpha}_n|_0,\forall n\}\leq{}\beta_1\\
G(\mc{C})     &      & otherwise
\end{array} \right.
\end{equation}
where $\beta_1$=0.1 is a predefined threshold to select coarse superpixels that are highly similar to false positives recorded in the flip dictionary. Similarly, the false negatives can be detected from the coarse superpixels with initial labels of 0 by using the flip dictionary $\bl{D}_i^{f-}$. Such a simple strategy can provide a rough refinement of superpixel labels so that the workload of the human annotator can be gradually reduced.

%
%\begin{figure*}[t]
%\centering
%\includegraphics[width=1.00\textwidth,height=8.5cm]{empty.png}
%\caption{Representative images in the synthetic dataset. Images marked with red borders indicate that the semantic objects are contained both in both ImageNet and MS-COCO, indicating that they can be annotated under strong supervision. Images marked with green borders do not have corresponding MS-COCO images with similar semantic objects (\eg, the xxxx in the xx row), indicating that these images need to be annotated with only weak and click supervisions.}
%\label{fig:repExample}
%\end{figure*

\section{Experiments}

\begin{figure*}%\label{fig:imageNetSample}
\includegraphics[width=1.00\textwidth]{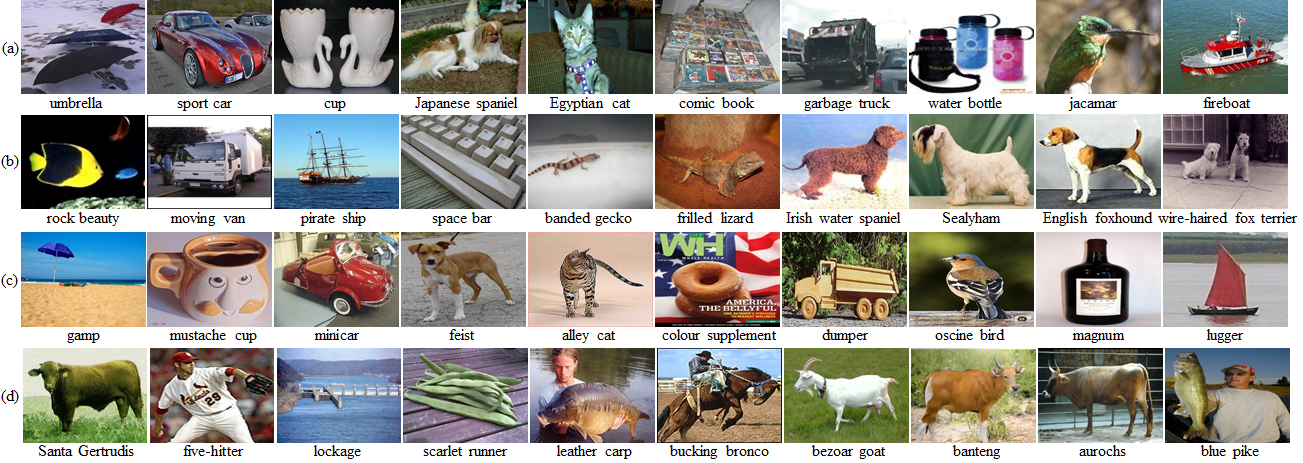}
\caption{Representative images from four groups. (a)~Group-$D$-$C$, (b)~Group-$D$-$\tilde{C}$, (c)~Group-$\tilde{D}$-$C$ and (d)~Group-$\tilde{D}$-$\tilde{C}$. By measuring the inter-group and intra-group performance variation, we can provide a comprehensive evaluation.}
\label{fig:imageNetSample}
\end{figure*}

\subsection{Settings}
The agent in our approach is dynamic and evolves along with the annotation process. To validate the effectiveness of such a dynamic agent, we build a synthetic dataset that consists of 120 image sets selected from:

\myPara{MS-COCO}~\cite{lin2014microsoft}. This dataset contains 328k images with objects annotated at the instance level. We crop a tight bounding box that covers each semantic object to generate a new image, which is also assigned the same tag with the semantic object. In total, we collect 80 image sets with 123,287 images. These images are used to simulate the tagged images with previously annotated per-pixel masks.

\myPara{ImageNet}~\cite{deng2009imagenet}. This dataset contains $21,841$ tagged image sets. From this dataset, we randomly select 20 image sets from the $1,000$ synsets that are previously used to train the DPN model~\cite{chen2017dual}, and another 20 image sets from the rest ones. From each set, we randomly select 10\% images and collect 2,389 images within 40 image sets that form four groups:
\begin{enumerate}
\item Group-$D$-$C$: 10 sets (1096 images) that are \bl{used} in training DPN and have \bl{the same} tags with the 80 MS-COCO image sets.
\item Group-$D$-$\tilde{C}$: 10 sets (574 images) that are \bl{used} in training DPN and have \bl{different} tags with the 80 MS-COCO image sets.
\item Group-$\tilde{D}$-$C$: 10 sets (659 images) that are \bl{not used} in training DPN and have \bl{the same} tags with the 80 MS-COCO image sets.
\item Group-$\tilde{D}$-$\tilde{C}$: 10 sets (510 images) that are  \bl{not used} in training DPN and have \bl{different} tags with the 80 MS-COCO image sets.
\end{enumerate}
The images from these four groups are used to simulate the tagged images whose per-pixel masks need to be annotated. Some representative images obtained from these four groups can be found in Fig.~\ref{fig:imageNetSample}. We find that images from Group-$D$-$C$ and Group-$\tilde{D}$-$C$ contain semantic objects that are also contained in certain images from MS-COCO, indicating that these images can be annotated under strong supervision. Other images from Group-$D$-$\tilde{C}$ and Group-$\tilde{D}$-$\tilde{C}$ do not have corresponding images in MS-COCO, indicating that these images need to be annotated mainly with weak and click supervisions. In this manner, we can explore the influence of various types of supervisions. In addition, the DPN model is trained to best extract representative features of primary objects from images of Group-$D$-$C$ and Group-$D$-$\tilde{C}$. As a result, the performance of our approach on these groups, when compared with those on  Group-$\tilde{D}$-$C$ and Group-$\tilde{D}$-$\tilde{C}$, can be used to depict the generalization ability of the proposed approach to new images.

Given the four groups, ten subjects (eight males and two females, aged from 20 to 28) are requested to annotate the semantic objects from the 40 image sets (four image sets per subject) by using the LabelMe toolbox~\cite{russell2008labelme}. In this manner, we obtain a binary object mask for each image from the 40 image sets. These binary masks are used as the ground-truth in all experiments. Let $\mc{M}_\mc{I}^0$ be the ground-truth mask of an image $\mc{I}$, we access the quality of a binary mask $\mc{M}_\mc{I}^1$ generated by another approach with the metrics of Precision and Recall. Correspondingly, the overall performance of an approach, a.k.a. the agreement with the masks generated by the LabelMe toolbox, is measured by the F-measure:
\begin{equation}
\text{F-measure}=\frac{2\cdot{}\text{Precision}\cdot{}\text{Recall}}{\text{Precision}+\text{Recall}}
\end{equation}

%\begin{figure*}[t]
%\centering
%\includegraphics[width=1.00\textwidth,height=11.0cm]{empty.png}
%\caption{Representative results generated by LabelMe and our approach.}
%\label{fig:qualityCompare}
%\end{figure*}

\subsection{Annotation Quality Test}
A successful annotation approach should provide satisfactory per-pixel masks while greatly reducing the time cost of the human annotator. To validate our approach from these two perspectives, we request eight additional subjects to annotate the semantic objects from all the 40 images sets with our annotation tool. The annotation results are compared with the masks generated by the ten subjects with LabelMe toolbox to show the annotation quality. Moreover, the flip dictionary is inhibited in the annotation process so that the quality of the initialization results provided by the agent indirectly reflects the speed up factor. An inherent assumption here is: the better the initialization results, the faster the manual annotation process.

\begin{figure*}
\includegraphics[width=1.00\textwidth]{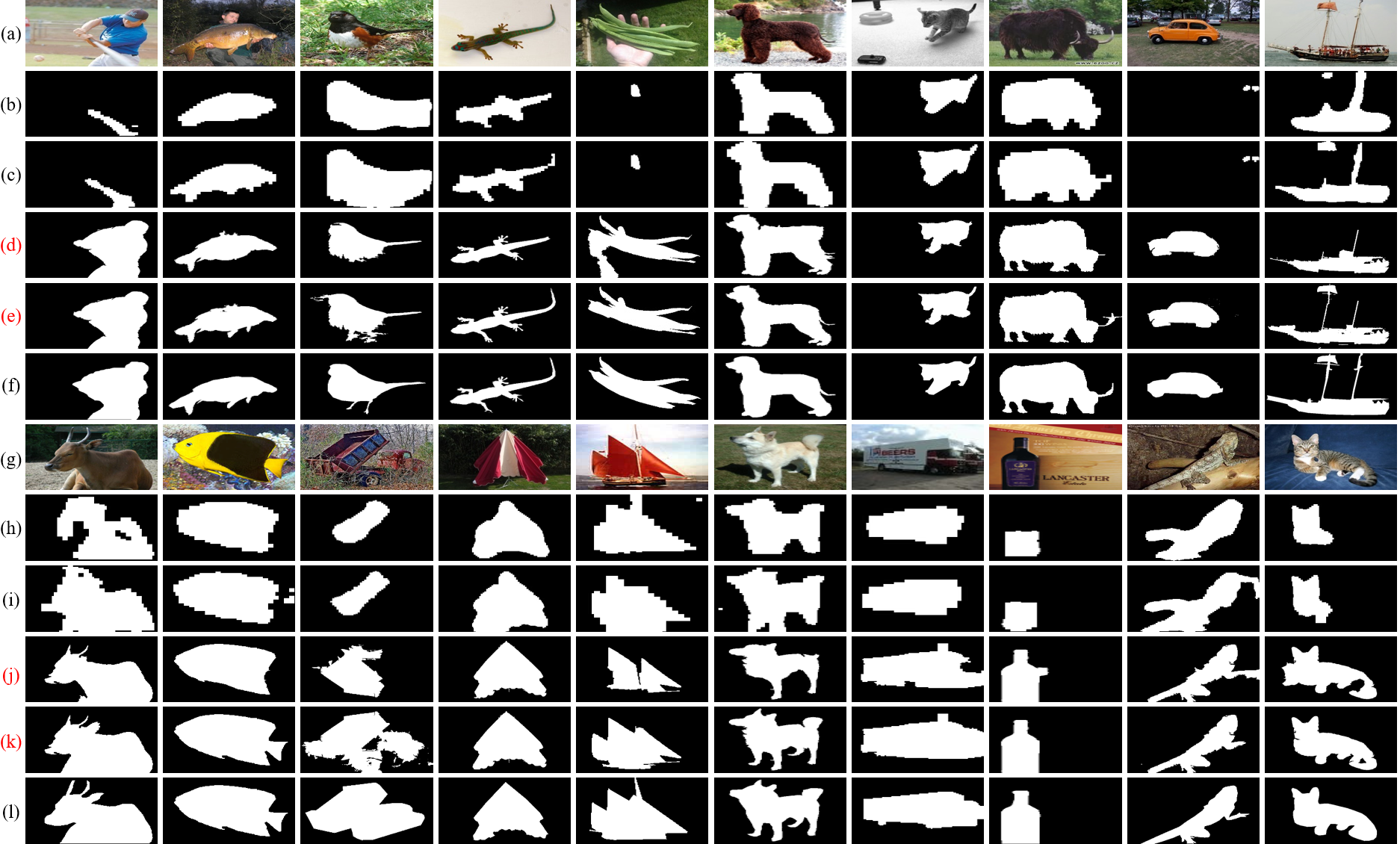}
\caption{Representative annotation results. (a,g) Tagged images, (b,h) DeepMask, (c,i) SharpMask, (d,j)~Our initialization results, (e,k)~Our final annotation results, (f,l)~LabelMe results (ground-truth).}
\label{fig:Final-result}
\end{figure*}

The performance of our automatic initialization and manual annotation results are shown in Table \ref{tab:performance}, and some representative results can be found in Fig.~\ref{fig:Final-result}. In Table \ref{tab:performance} and Fig.~\ref{fig:Final-result}, we also demonstrate the results from two state-of-the-art models, DeepMask~\cite{pinheiro2015learning} and SharpMask~\cite{pinheiro2016learning}, to provide an intuitive comparison with our initialization results. For an input image, DeepMask uses a CNN model to output a pixel labeling of an object. SharpMask provides a refinement module to obtain more accurate results. Note that both DeepMask and SharpMask are trained on the MS-COCO dataset and we use their top-1 proposals as the labeling results.

\begin{table}[t]
\centering{
\caption{Annotation qualities (in \%) of four approaches on the 40 image sets. The top two approaches are highlighted with bold and underline, respectively.} \label{tab:performance}
\begin{tabular}{cccc} \toprule
Model   &Precision &Recall  &F-Measure \tabularnewline \midrule
Random        &31.36  &49.96  &38.53  \tabularnewline
DeepMask      &52.13  &62.34  &56.78  \tabularnewline
SharpMask     &54.34  &61.71  &57.79  \tabularnewline
Our-Init      &\ul{60.35}  &\ul{72.42}  &\ul{65.83}  \tabularnewline
Our-Final     &\bl{91.33}  &\bl{91.09}  &\bl{91.21}  \tabularnewline
\bottomrule
\end{tabular}
}
\end{table}

\begin{table}[t]
\centering{
\caption{The best and worse image sets in terms of annotation qualities (in \%) of our approach.} \label{tab:qualityRank}
\begin{tabular}{c@{}cccc} \toprule
    &Categories  &~Precision &Recall  & F-Measure \tabularnewline \midrule
\bl{\multirow{5}{*}{\rotatebox{90}{Top 5}}}
    &~colour supplement  &99.11  &94.83  &96.93  \tabularnewline
    &~comic book         &96.57  &95.78  &96.17  \tabularnewline
    &~Irish water spaniel&94.79  &97.30  &96.03  \tabularnewline
    &~mustache cup       &95.15  &96.11  &95.63  \tabularnewline
    &~Egyptian cat       &94.31  &96.36  &95.32  \tabularnewline \hline
\bl{\multirow{5}{*}{\rotatebox{90}{Bottom 5}}}
    &~banded gecko       &88.69  &84.90  &86.75\tabularnewline
    &~bucking bronco     &83.86  &85.35  &84.60\tabularnewline
    &~lockage            &92.17  &75.71  &83.14\tabularnewline
    &~pirate ship&75.87  &85.31 &80.31\tabularnewline
    &~fireboat           &77.67  &74.25 &75.93\tabularnewline
\bottomrule
\end{tabular}
}
\end{table}

From Table~\ref{tab:performance}, we can see that the agreement between the annotation results of our approach and LabelMe reaches up to 91.23\% in terms of F-measure. To further investigate this, we demonstrate in Table~\ref{tab:qualityRank} the top five and bottom five image sets selected according to the F-measure of our approach.

For the top 5 image sets such as comic book and mustache cup, the annotation results from two annotation approaches perfectly match each other. Even for the challenging image sets such as Irish water spaniel, we still have an agreement above 96.03\%. Interestingly, we can see that the in the top 5 image sets, most of them have corresponding semantic objects in MS-COCO (except Irish water spaniel), indicating that the strong supervision plays an important role in improving the annotation quality. In contrast, the bottom 5 categories usually have no such strong supervision (except fireboat). Moreover, the objects in the bottom five sets may have diversified appearance, which may prevent them from being easily annotated and may lead to annotation ambiguity even between different subjects.

Beyond the annotation quality, we find that the time cost of the annotator may also be remarkably reduced due to the impressive performance of the computer agent in initializing the object masks. On average, the F-measure of our initialization results reaches up to 65.83\%, which outperforms DeepMask (56.78\%) and SharpMask (57.79\%) by 9.05\% and 8.04\%, respectively. Note that the time-consuming pre-training processes are not required by our approach, in which the agent is dynamically generated by retrieving related images and objects. This implies that our approach can provide better initialization results and can be efficiently deployed for annotating various types of semantic objects.

In addition, we compare the initialization results with the final annotation results with our approach and find that only 25.02 coarse superpixels per image have wrong labels or inaccurate boundaries that need to be flipped or divided through left or right mouse clicks. Considering that such mouse click are much easier to conduct than dragging and dropping the anchor points, we can safely claim that the proposed approach can greatly reduce the time cost of human-being in annotating semantic objects.

\subsection{Performance Analysis}
We further analyze the performance of our approach from the perspective of generalization ability and agent evolvability. To verify the generalization ability, we list in Table~\ref{tab:compare} the performance of our initialization results on the four groups. From Table~\ref{tab:compare}, we can find that our annotation approach have comparable performance for various types of images inside or outside the training set of DPN and the MS-COCO dataset, implying an impressive generalization ability.

%In addition, it can also adapt to completely new categories such as book, umbrella  and cup, while many previous models such as Polygon-RNN~\cite{castrejon2017annotating} may have difficulties to handle with if the training data do not contain such categories.

\begin{figure}[t]
\includegraphics[width=1.00\columnwidth,height=3cm]{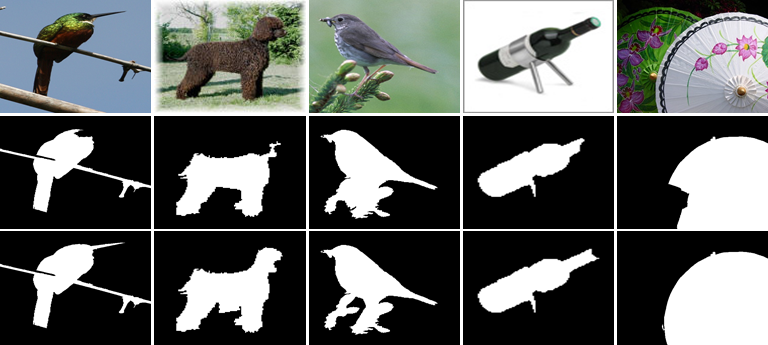}
\caption{Effects of automatic annotation refinement with click supervision. The second row shows the initialization results, and the third row shows the object masks after automatic refinement with the flip dictionaries.}
\label{fig:refine}
\end{figure}

To validate the evolvability of the agent, we equally divide each of the 40 image sets into two click collection sets and a click verification set. On the click collection sets, we collect the mouse clicks of annotators to form a flip dictionary, and compare the agent performance over the click verification sets with or without using the flip dictionaries. We find that by using the flip dictionaries formed on zero, one or two click collection sets, the agent obtains a F-measure of 64.53\%, 65.14\% and 65.27\% on the images from the click verification set, respectively. Note that for each of the 40 image sets, the flip dictionaries are formed only on hundreds of clicks (\ie, 59.72 images per image set and 25.02 coarse superpixels get manually clicked per image). Some representative results before and after the automatic agent refinement with the flip dictionaries can be found in Fig.~\ref{fig:refine}. These results indicate that the propose approach has the capability of memorizing past failures and can gradually evolve to avoid them when facing similar images. In this manner, the agent gradually becomes smarter and thus save more time of the annotator.

\begin{table}[t]
\centering{
\caption{Agent performances (in \%) over four groups with only weak and strong supervisions} \label{tab:compare}
\begin{tabular}{cccc} \toprule
Group                           &Precision  & Recall & F-Measure  \tabularnewline   \midrule
Group-$D$-$C$                   &64.76      &67.42   &66.06       \tabularnewline
Group-$D$-$\tilde{C}$           &56.98      &76.80   &65.42       \tabularnewline
Group-$\tilde{D}$-$C$           &61.02      &71.80   &65.97       \tabularnewline
Group-$\tilde{D}$-$\tilde{C}$   &53.78      &79.03   &64.00       \tabularnewline
\bottomrule
\end{tabular}
}
\end{table}

%In the evolvability test, we select 10 new images from each of the top five and bottom five image sets (see Table~\ref{tab:qualityandspeed} for the tags of these ten sets). After that, we request 20 subjects to annotate all the newly selected 100 images. In the annotation process of ten subjects, the Flip  dictionary constructed previously are used, while the other ten subjects adopt empty Flip and Divide dictionaries. On average the former ten subjects spend xx.xs on average in handling an image, while the latter ten spend xx.xs on average.

%
%\begin{figure}[t]
%\centering
%\includegraphics[width=1.00\columnwidth,height=5.0cm]{empty.png}
%\caption{IoU scores between the initialized and final annotation results when different Weak dictionary sizes are used.}
%\label{fig:weakSize}
%\end{figure}

\section{Conclusion}
Per-pixel masks of semantic image objects are difficult to be obtained due to the tedious and inefficient annotation process. In this paper, we propose a human-agent collaborative annotation approach that aims to convert existing large-scale datasets of tagged images to per-pixel masks of semantic objects. Multi-granularity supervisions from images with similar tags, previously annotated objects and annotator clicked superpixels are incorporated into the collaborative framework in the form of weak, strong and click supervisions. Compared with previous annotation tools, the proposed agent is dynamically formed and can gradually evolve to recover failures. Experimental results on a synthetic dataset show that the proposed approach performs impressively on annotating tagged images. We believe this approach can be help to the construction of large-scale image datasets with per-pixel masks of semantic objects.

%\begin{acks}

%\end{acks}

\bibliographystyle{ACM-Reference-Format}
\bibliography{sample-bibliography}
%\bibliography{egbib}

\end{document}